\documentclass[letterpaper, 10 pt, conference]{ieeeconf}  

\IEEEoverridecommandlockouts                              

\usepackage{times}
\usepackage{epsfig}
\usepackage{graphicx}
\usepackage{amsmath}
\usepackage{amssymb}

\usepackage{color}
\usepackage{multirow}
\usepackage{multicol}
\usepackage{bm}
\usepackage[pagebackref=true,breaklinks=true,letterpaper=true,colorlinks,bookmarks=false]{hyperref}
\usepackage{caption}
\usepackage{comment}

\newcommand{\tref}[1]{Tabel.~\ref{#1}}
\newcommand{\fref}[1]{Fig.~\ref{#1}}
\newcommand{\etal}{{\em et al.}}
\newcommand{\eg}{\textit{e}.\textit{g}.}

\title{\LARGE \bf
SegVoxelNet: Exploring Semantic Context and Depth-aware Features for 3D Vehicle Detection from Point Cloud
}

\author{Hongwei Yi$^{1*}$\thanks{*Corresponding author: {\tt\small hongweiyi@pku.edu.cn}}, Shaoshuai Shi$^{2}$, Mingyu Ding$^{3}$, \\ Jiankai Sun$^{2}$, Kui Xu$^{4}$, Hui Zhou$^{5}$, Zhe Wang$^{5}$, Sheng Li$^{6}$, Guoping Wang$^{6}$ \\
\vspace{-15pt}
\thanks{$^{1}$Shenzhen Graduate School, Peking University  $^{2}$The Chinese University of Hong Kong  $^{3}$The University of Hong Kong $^{4}$Tsinghua University  $^{5}$SenseTime Research  $^{6}$School of EECS, Peking University}
}

\begin{document}

\maketitle
\thispagestyle{empty}
\pagestyle{empty}

\begin{abstract}
3D vehicle detection based on point cloud is a challenging task in real-world applications such as autonomous driving. Despite significant progress has been made, we observe two aspects to be further improved. First, the semantic context information in LiDAR is seldom explored in previous works, which may help identify ambiguous vehicles. Second, the distribution of point cloud on vehicles varies continuously with increasing depths, which may not be well modeled by a single model. In this work, we propose a unified model SegVoxelNet to address the above two problems. A semantic context encoder is proposed to leverage the free-of-charge semantic segmentation masks in the bird's eye view. Suspicious regions could be highlighted while noisy regions are suppressed by this module. To better deal with vehicles at different depths, a novel depth-aware head is designed to explicitly model the distribution differences and each part of the depth-aware head is made to focus on its own target detection range.
Extensive experiments on the KITTI dataset show that the proposed method outperforms the state-of-the-art alternatives in both accuracy and efficiency with point cloud as input only.
\end{abstract}

%
%

%


%

%
\maketitle

\section{Introduction}
Deep neural networks (DNN) have made tremendous advances on the task of 2D visions such as vehicles and other objects detection~\cite{liu2016ssd, faster-rcnn} and semantic segmentation~\cite{he2017maskrcnn, yang2018segstereo}. However, compared with its image-based counterparts, 3D scene understanding is still under-explored while crucial for many real-world applications such as autonomous driving. In autonomous driving, LiDAR is one of the most important sensors, which captures the 3D structure of the scene by sparse point cloud. In this work, we focus on 3D vehicle detection, whose goal is to localize the 3D oriented bounding boxes of physical vehicles from the point cloud.


\begin{figure}
    \centering
    \begin{tabular}{cc}
        \hspace{-0.03\columnwidth}\raisebox{15mm}{\footnotesize{\textbf{a)}}} & \hspace{-0.04\columnwidth}\includegraphics[width=0.93\columnwidth]{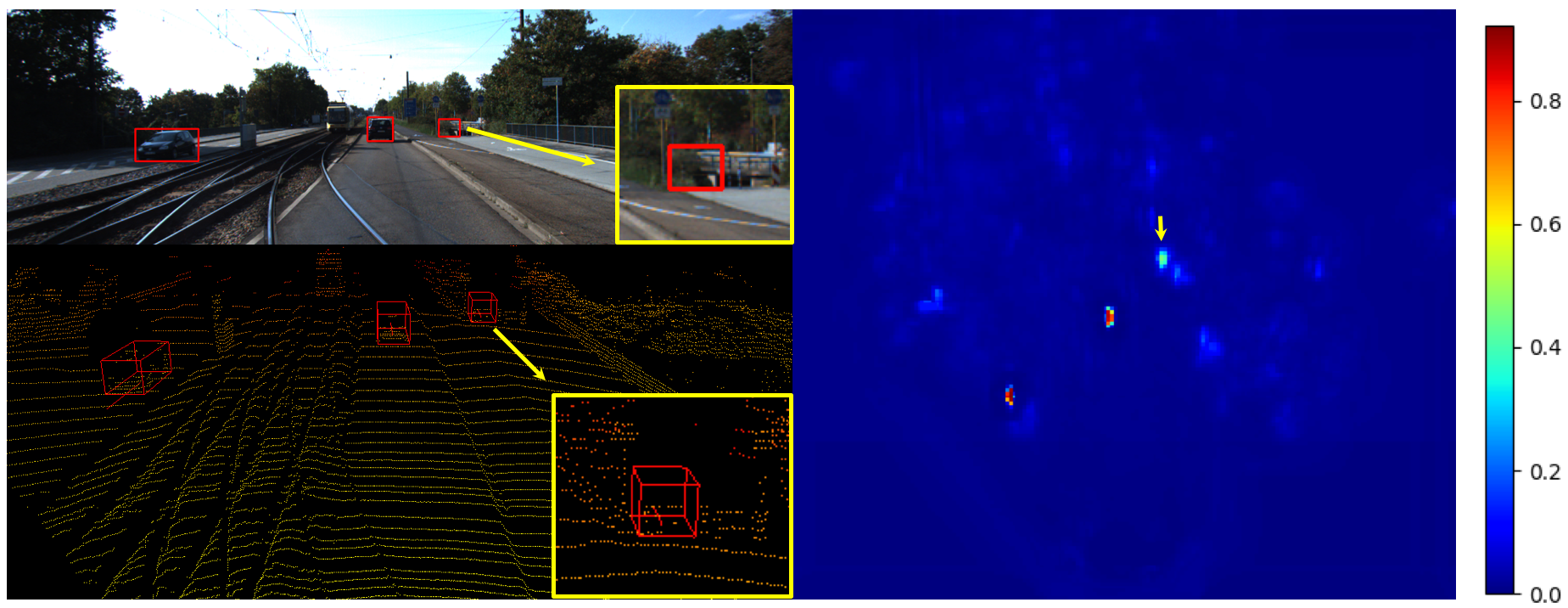} \\
        \vspace{-3pt}
        \hspace{-0.03\columnwidth}\raisebox{25mm}{\footnotesize{\textbf{b)}}} & \hspace{-0.04\columnwidth}\includegraphics[width=0.93\columnwidth]{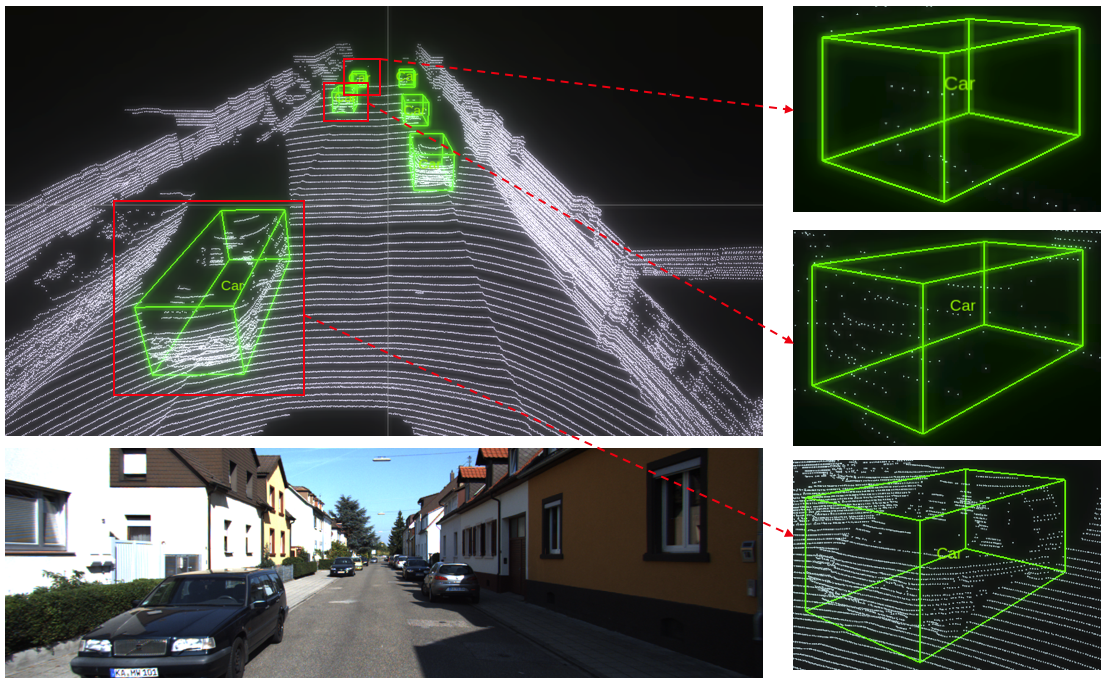} \\
    \end{tabular}
    \vspace{-3pt}
    \caption{a) An example of false positives in SECOND~\cite{yan2018second}. The left are the input image and LiDAR point cloud, the predict 3D vehicle results are marked by red boxes in the image and point cloud, respectively. The right is the class score map of range $ (0, 1)$ derived from its detection head. Yellow arrows highlight the false positive example due to local ambiguity outside the road. b) Visualization of the distribution of point clouds of vehicles with different depth.}
    \label{intro} 
    \vspace{-6pt}
\end{figure}

By leveraging the success in vehicle detection in 2D images, 3D vehicle detectors that combine both image and point cloud have been proposed.~\cite{qi2018frustum} first detects the 2D bounding boxes in the image and then detects 3D boxes in the point cloud constrained in the frustum.~\cite{chen2017multi, ku2018joint} encode the image as additional features and enrich the point cloud representation using a 2D detection backbone. However, such methods rely on strict time synchronization and extrinsic parameter calibration between LiDAR and the camera, which may not be satisfied in practical systems.

Methods exploiting point cloud only as input have also been explored in many works. 
Similar to 2D detection, these methods can be mainly divided into two categories: the one-stage method and the two-stage method.
Generally, one-stage methods are faster than two-stage ones. 
Usually, these methods  encode the point cloud as a bird's eye view  (BEV)~\cite{chen2017multi, yang2018pixor} representation or to the irregular pillars~\cite{lang2018pointpillars}, and directly predicts the 3d bounding boxes and their scores. ~\cite{voxelnet, yan2018second} firstly group point cloud into voxels and then extracts features using 3D convolution from voxels which is used in region proposal network  (RPN)~\cite{girshick2015fast}.
The two-stage method usually generates bounding box proposals in the first stage and refine them in the second stage~\cite{himmelsbach2008LiDAR}. Recently,~\cite{shi2019pointrcnn} achieves impressive 3D detection results on the KITTI~\cite{Geiger2012CVPR} benchmark by introducing~\cite{qi2017pointnet++} as a two-stage encoder for canonical 3D bounding box refinement. 
However, these methods often need a separate model for each stage which is time-consuming.  

Although these methods have made impressive progress, we argue that two main problems still remain unsolved. First, it is difficult for current models to distinguish ambiguous vehicles without semantic context information. As shown in \fref{intro} a), our baseline method detects a false positive caused by an ambiguous obstacle outside the road. 
Second, the density distribution of point cloud varies continuously for vehicles at different depths, as shown in \fref{intro} b). It  may be difficult for the network to simultaneously model the distribution across different depths.

Based on the above observation, we design a unified model called SegVoxelNet by exploiting free-of-charge bird's eye view (BEV) semantic masks as additional supervision signal and a depth-aware head for learning distinctive depth-aware features for vehicles at various depths. 
The whole network is fully-convolutional and end-to-end trainable. Experiments on the well-known KITTI~\cite{Geiger2012CVPR} dataset show that our proposed method achieves considerable improvement over state-of-the-art methods.
  
Our main contributions are summarized below:
\begin{itemize}
\item A unified framework called SegVoxelNet that incorporates free-of-charge semantic segmentation information into 3D vehicle detection pipeline, where semantic context provides guidance for 3D vehicle detection.
\item A depth-aware head with convolutional layers of different kernel sizes and dilated rates that improve feature learning of vehicles at different depths. 

\item Our SegVoxelNet achieves state-of-the-art results on the KITTI dataset with real-time efficiency.
\end{itemize}

\begin{figure*}[t]
  \centering
  \includegraphics[width=1.9\columnwidth]{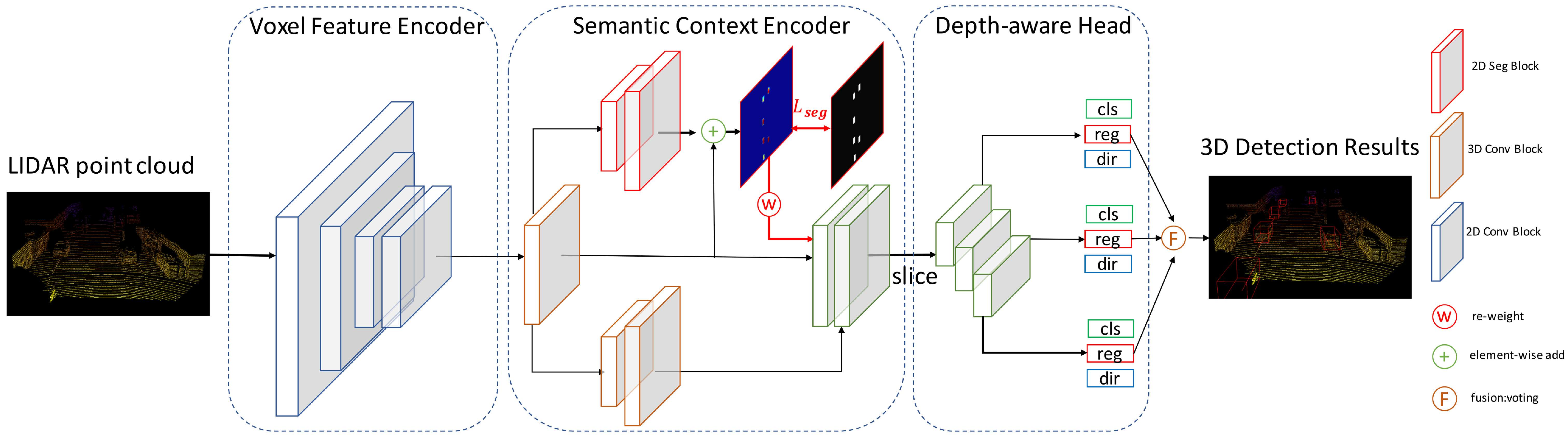}
  \vspace{-3pt}
  \caption{An overview of the proposed network SegVoxelNet, which consists of a Voxel Feature Encoder (VFE),  a Semantic Context Encoder (SCE) and a depth-aware head. 
  VFE encodes the raw point cloud into voxels and further converts them into bird's eye view feature maps. SCE predicts the semantic segmentation masks and encodes the semantic context information in the feature maps for better detection. And the depth-aware head is designed to have multiple parts to predict 3D bounding boxes for vehicles in different depth level.}
  \label{architecture}
  \vspace{-17pt}
\end{figure*}

\section{Related Work}
In this section, we briefly review the existing methods for 3D vehicle detection, which could be classified into the following three categories based on different input settings.

\textbf{3D Vehicle Detection from Images: }
Some works directly predict the 3D bounding boxes based on the RGB images. Chen \etal\cite{chen2016monocular} encoded the 3D locations, context and shape features of 3D vehicles as the energy function to score the exhaustively placed 3D bounding boxes on the estimated ground plane. Mousavian \etal\cite{mousavian20173d} recovered the 3D locations by leveraging the geometry constraints between 2D and 3D bounding boxes and predicted the orientations of vehicles by their proposed MultiBin architecture.~\cite{chabot2017deep, manhardt2018roi} predicted the 3D bounding boxes by estimating the 3D vehicle shape from the introduced CAD models.  Some works~\cite{chen20183d, licvpr2019, wangcvpr2019} further utilize stereo images for better estimating the depth information to detect 3D vehicles. However, no matter using monocular or stereo images, there is still a large performance gap between these methods and point cloud based methods, as the vehicle depth estimated from images is far from being accurate.

\textbf{3D Vehicle Detection from Multiple Sensors: }
Some methods explored to fuse information from the point cloud and RGB image to improve the performance of 3D vehicle detection. Chen \etal\cite{chen2017multi} and Ku \etal\cite{ku2018joint} encoded the point cloud as bird's eye view feature maps and projected the 3D proposals to different views (\eg bird's eye view for point cloud and front view for image) to crop object features from different sensors for the final 3D bounding box prediction.
 Qi \etal\cite{qi2018frustum} and Xu \etal\cite{xu2018pointfusion} exploited the 2D image detectors to generate 2D proposals from images, which are then used to crop the point cloud within each 2D box for the following 3D box estimation by applying PointNet~\cite{qi2017pointnet, qi2017pointnet++} to these point cloud. However, currently the bird's eye view based methods suffer from information loss of quantization and feature misalignment problem of different sensors, and the 2D image proposal based methods heavily depend on the performance of 2D detectors and may fail on the occluded objects. Unlike these methods, our method directly predicts 3D bounding boxes from the 3D space generated by the point cloud, which is both efficient and natural to process the occluded objects since they are separated in the 3D space.

\textbf{3D Vehicle Detection from Sparse Point Cloud: }
Detecting 3D vehicles directly from the raw point cloud is practical and important for autonomous driving since it avoids the sensor synchronization problem. Zhou \etal\cite{voxelnet} first proposed the VoxelNet architecture to predict 3D bounding boxes with point cloud, and the following work~\cite{yan2018second} combined VoxelNet with sparse convolutions~\cite{3DSemanticSegmentationWithSubmanifoldSparseConvNet} to further improve efficiency and effectiveness. Other several works~\cite{yang2018pixor, lang2018pointpillars} project the point cloud to bird's eye view and utilizing 2D CNN to predict the 3D bounding boxes from bird's view feature maps.
Shi \etal\cite{shi2019pointrcnn} proposed the PointRCNN architecture to directly generate 3D proposals from raw point cloud by segmenting the foreground points and refine them in the canonical coordinates.
Different from these methods, our proposed one-stage detector SegVoxelNet further explores the semantic context information as a guidance to benefit the confidence prediction and 3D box generation, and the new depth-aware detection architecture also improves the 3D detection performance by learning separate features for point cloud at different depths.

\section{Method} \label{method}
In this section, we introduce our proposed single stage detection framework SegVoxelNet, as illustrated in \fref{architecture}. 

\subsection{Voxel Feature Encoder}

The Voxel Feature Encoder (VFE) is applied to the raw point cloud to obtain voxelized feature representation for the following SCE module. It consists of two steps, i.e., \textit{point cloud voxelization} and \textit{voxel feature extraction}. 

We first partition the point cloud into equally spaced voxels $v$ with the size $ (v_W, v_L, v_H)$. 
To save up computation and reduce the imbalance for number of points between voxels, a fixed number of 3D points are randomly sampled in each voxel. Finally, the averaged coordinates of the 3D points $(\bar{x}, \bar{y}, \bar{z})$ in the voxel is taken as the feature for that voxel. 
The voxelized point cloud is then processed by four repeating $Block$s sequentially. Each $Block$ consists of several 3D sub-manifold sparse convolution layers and a normal sparse convolution layer for down-sampling in x,y-axis and squeezing z-axis. After each sparse convolution layer, BatchNorm layer and ReLU layer are applied. 
Finally, the output feature maps of the last $Block$ in the VFE are reshaped into 2D BEV format tensors for further analysis.

\subsection{Semantic Context Encoder} 

The Semantic Context Encoder (SCE) in \fref{architecture} takes the BEV feature maps from the Voxel Feature Encoder as input, and output the semantic context encoded feature maps for detection. 
The proposed SCE consists of two branches sharing the input VFE feature maps. The first branch is modified from SECOND~\cite{yan2018second}. 
It consists of a U-Net structure with a downsampling and an upsampling layer for obtaining larger receptive field.  The output of the U-Net is the same size as the input, so the it can be concatenated to the input VFE feature map to generate the main feature map.
The second branch learns to predict the BEV semantic masks, which is then used to enhance the feature maps from the first branch by a fusion module. 

For the semantic segmentation branch, the semantic masks can be easily calculated by projecting the 3D ground truth boxes to BEV. Here we explore two types of masks:

\begin{itemize}
    \item voxel-type mask. We first match the 3D ground truth boxes to voxels. Only non-empty voxels which overlap with the boxes will be assigned as foreground. 
    \item box-type mask. We directly project the 3D ground truth boxes to BEV, and set voxels in the boxes to foreground.
\end{itemize}
Both types of masks can be used to train the SCE. 

Inspired by FPN~\cite{FPN}, the semantic segmentation branch consists of a feature pyramid to extract multi-scale context information from the BEV image. Specifically, The network consists of $5$ residual blocks, $2$ maxpooling layers, $2$ upsampling layers and $2$ convolutional layers followed by BatchNorm and ReLU layers for fusing multi-scale feature maps.
The segmentation branch can be trained by the commonly used cross entropy loss or softmax loss.

Then, we fuse the two branches to obtain the semantic context encoded feature maps for the detection head. The fusion can be done by re-weighting the feature maps by the probability map as attention residual learning in~\cite{wang2017residual},  Formally, given the probability map $M$ and the feature map of the first branch $F$, the re-weighted feature $R$ can be calculated by
\begin{equation}
    R_c(x,y)= (1+M(x,y)) \cdot F_c(x,y)
\end{equation}
where $c$ indexes the channel, $x,y$ index the spatial location and M(x,y) is the probability map from segmentation branch ranging from $[0,1]$.
This formulation enhances car related features and suppresses noises from trunk features.

\begin{figure*}
  \centering
  \begin{tabular}{cc}
    \hspace{+0.01\columnwidth}\includegraphics[width=0.99\columnwidth]{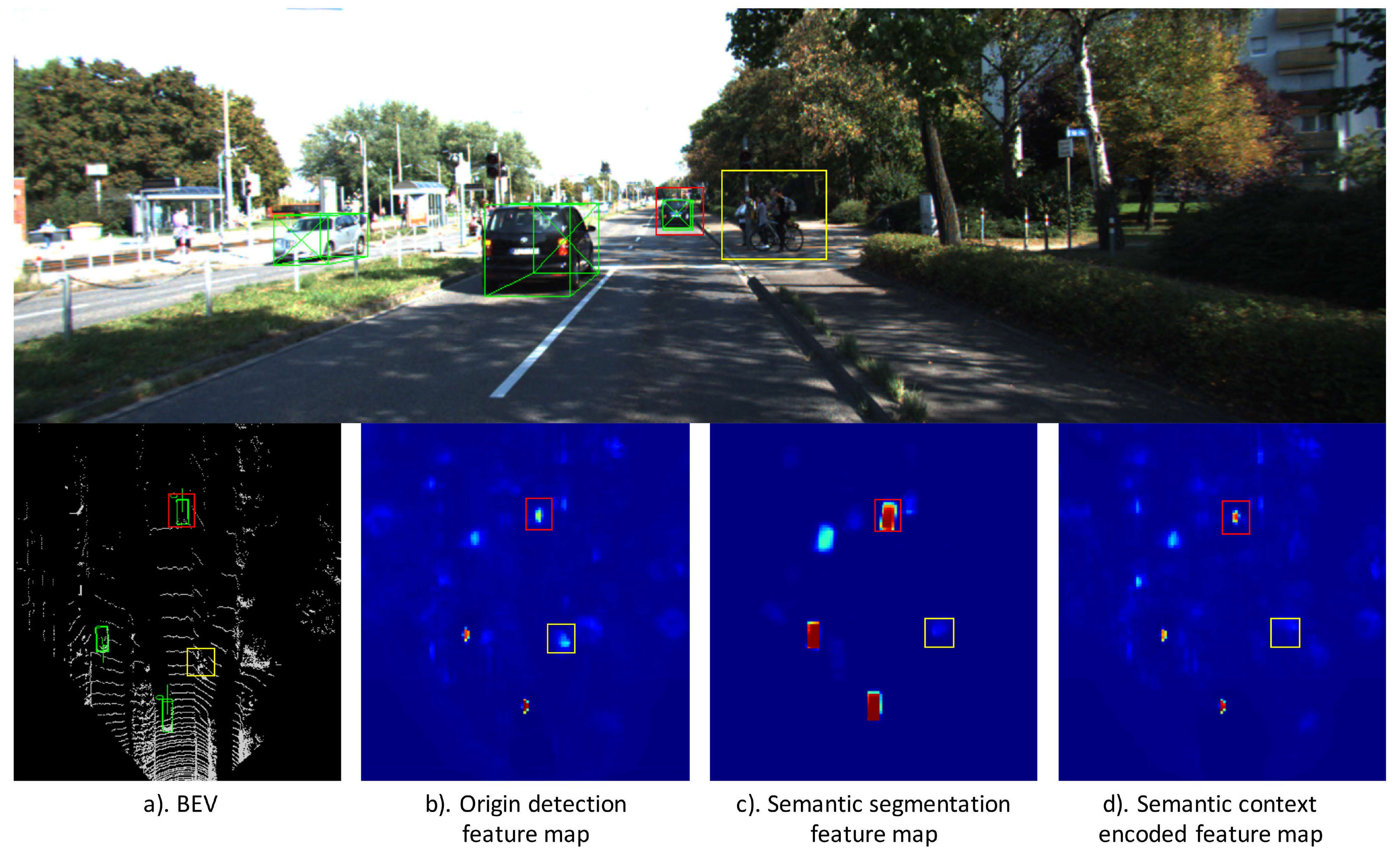} &
    \hspace{-0.04\columnwidth}\includegraphics[width=0.99\columnwidth]{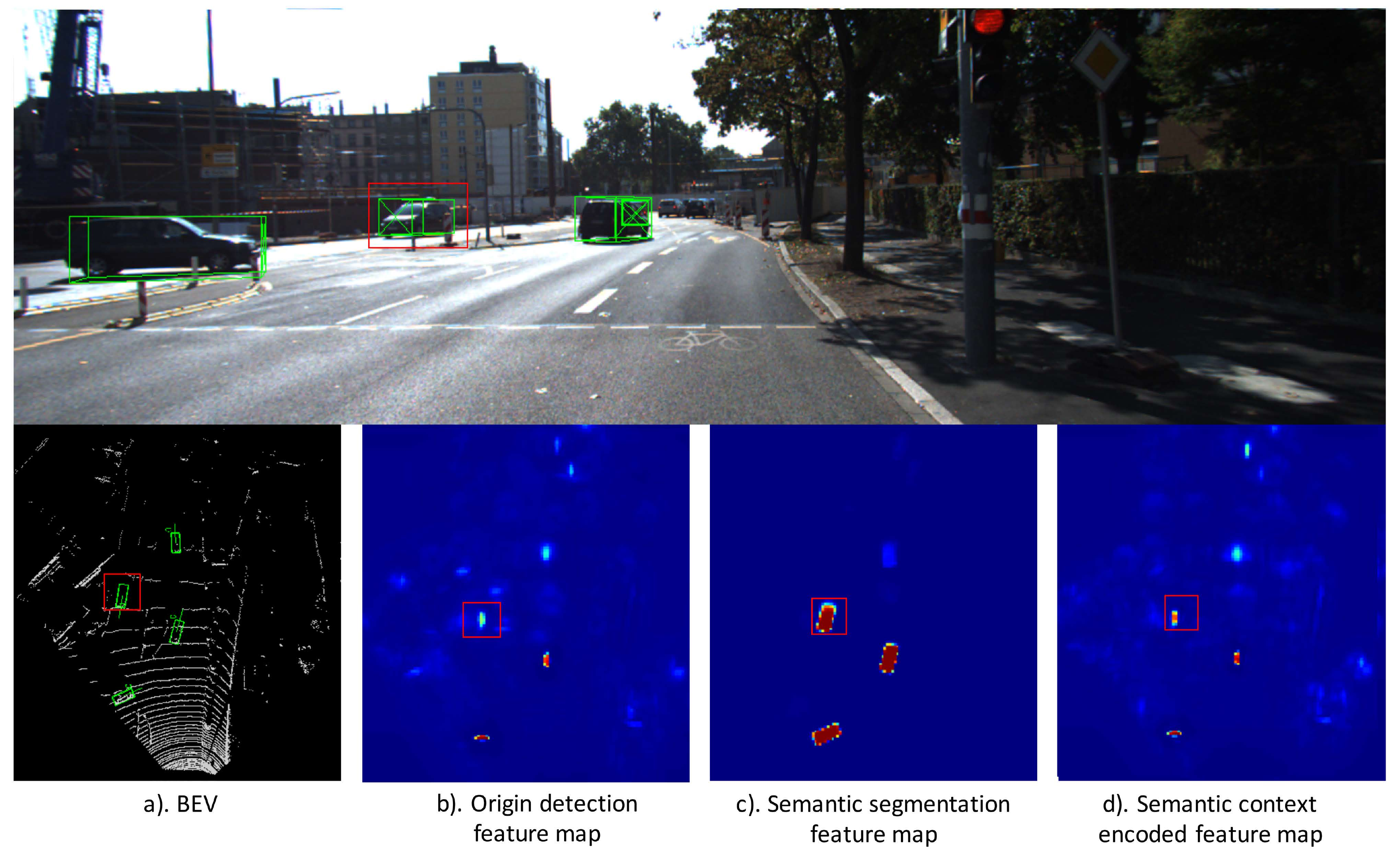}
  \end{tabular}
  \vspace{-7pt}
  \captionof{figure}{Visualization of the influence of integrating semantic segmentation information into detection feature maps. Top is the 2D image, bottom consists of a) bird's eye view LiDAR point clouds, b) origin detection feature map, c) semantic segmentation feature map and d) semantic context encoded feature map. Green bounding boxes in the image and BEV denote the ground truth 3D vehicles. The red bounding boxes on the semantic segmentation feature map enhances the car-existing region on the detection feature map, while the yellow ones gives a suppressive effect on it.}
  \vspace{-17pt}
  \label{segRPN}
\end{figure*}

\subsection{Depth-aware Head} \label{detectionHead}
In 2D detector SSD~\cite{liu2016ssd}, the detection head receives different scales of RPN feature maps for classifying and regressing different sizes of vehicles. Different from the camera, Lidar is an active sensor, which scans the scene to generate point cloud. The sizes of vehicles are invariant with the depth in the point cloud. The scale invariance of vehicles in the point cloud cancels the necessity for multi-scale feature learning in the detection head. However, the density of point cloud on vehicles at different depths is quite different as demonstrated in \fref{intro}.  Generally, the number of points on a vehicle decreases rapidly with the increasing depth.  Based on these observations, we design a depth-aware head with convolution layers of different kernel sizes and dilated rates as illustrated in \fref{architecture}. 
To simultaneously increase the model capacity while keep the network as efficient as possible, we divide the input feature maps from SCE into three parts along the x-axis in LiDAR coordinate with two overlapping regions. The length of the overlapping region is designed as twice of the vehicle length. For the three parts, convolution layers with kernel size $1/3/3$ and dilated rate $1/1/2$ are used for learning distinctive features from three different distances, respectively. Each part is asked to solve the same detection multi-task, which can be trained by minimizing the same loss function in SECOND~\cite{yan2018second} $L_{D}$:
\begin{equation}
L_{D} = \lambda_{loc}L_{loc} + L_{cls} + \lambda_{dir}L_{dir}, 
\end{equation}
which is normalized by positive anchors and $\lambda_{loc}=2$, $\lambda_{dir}=0.2$.

To perform 3D vehicle detection,
ground truth boxes and anchors are defined as $ (x, y, z, w, l, h, \theta)$. The localization regression residuals between the ground truth and anchors are defined by:
\vspace{-1pt}
\begin{equation}
    \begin{aligned}
    {\Delta}x&=\frac{x_{gt}-x_{a}}{d_{a}}, & {\Delta}y&=\frac{y_{gt}-y_{a}}{d_{a}}, & {\Delta}z&=\frac{z_{gt}-z_{a}}{h_{a}}, \\
    {\Delta}w&=\log (\frac{w_{gt}}{w_{a}}), & {\Delta}l&=\log (\frac{l_{gt}}{l_{a}}), & {\Delta}h&=\log (\frac{h_{gt}}{h_{a}}),\\
    {\Delta}\theta&=\theta_{gt}-\theta_{a}, 
    \end{aligned}
\end{equation}
where $x_{gt}$ and $x_{a}$ are the ground truth and anchor boxes respectively, and $d_{a}=\sqrt{ (w_{a})^2+ (l_{a})^2}$. The localization loss function is:

\begin{equation}
    L_{loc}=\sum_{\omega\in (x,y,z,w,l,h,\theta)} \text{Smooth}L1 ({\Delta}\omega), 
\end{equation}

Since this loss can not distinguish the opposite orientation of the bounding box, we use a cross entropy loss on the classification of discretized directions~\cite{yan2018second}. For the vehicle classification, we use the focal loss~\cite{lin2017focal}.

Finally, We fuse the class score feature maps from three parts and choose the highest score of the class at each position. The detected vehicles are obtained by applying oriented NMS with bird's eye view IoU threshold $0.05$ to remove the overlapping bounding boxes.


\subsection{Final Loss Function}
The final loss function $L$ for training our proposed SegVoxelNet is defined as:
\begin{equation}
    L=\lambda_{S}L_{S}+\sum_{p=1}^{3}L_{D_{p}},
\end{equation}
where $p$ denotes the part index of depth-aware head. $\lambda_{S}$ is used to balance the weights for semantic segmentation and classification constrains. We set $\lambda_{S}=0.5$.
\begin{table*}[t]
    \centering
    \small
    \begin{tabular}{|c|c|c|c c c|ccc|ccc|}
    \hline
    \multirow{2}{*}{Method} & \multirow{2}{*}{Modality} & \multirow{2}{*}{Speed \tiny{(FPS)}} &  \multicolumn{3}{c|}{3D Box} & \multicolumn{3}{c|}{bird's Eye View} & \multicolumn{3}{c|}{Orientation} \\ \cline{4-12}
    & & & Easy & \underline{\textbf{Mod.}} & Hard & Easy & \underline{\textbf{Mod.}} & Hard & Easy & \underline{\textbf{Mod.}} & Hard \\ \hline
   \textbf{AVOD-FPN}~\cite{ku2018joint} & LiDAR \& RGB & 10 & 81.94 & 71.88 & 66.38 & 88.53 & 83.79 & 77.90 & 89.95 & 87.13 & 79.74 \\ \hline
   \textbf{F-PointNet}~\cite{qi2018frustum} & LiDAR \& RGB & 5.9 & 81.20 & 70.39 & 62.19 & 88.70 & 84.00 & 75.33 & - & - & - \\ \hline
   \textbf{UberATG-MMF}~\cite{Liang_2019_CVPR} & LiDAR \& RGB & 12 & 86.81 & 76.75 & 68.41 & 89.49 & 87.47 & 79.10 & - & - & - \\ 
   \hline \hline
   \textbf{PointRCNN}~\cite{shi2019pointrcnn} & LiDAR \tiny{(Two Stage)} & 10 & \textbf{84.32} & 75.42 & 67.86 & \textbf{89.28} & 86.04 & 79.02 & \textbf{90.76} & \textbf{89.55} & 80.76 \\ \hline 
   \textbf{VoxelNet}~\cite{voxelnet} & LiDAR & 4.4 & 77.47 & 65.11 & 57.73 & 89.35 & 79.26 & 77.39 & - & - & - \\ \hline
   \textbf{SECOND}~\cite{yan2018second} & LiDAR & 20 & 83.13 & 73.66 & 66.20 & 88.07 & 79.37 & 77.95 & 87.84 & 81.31 & 71.95 \\ \hline
   \textbf{PointPillars}~\cite{lang2018pointpillars} & LiDAR & \textbf{62} & 79.05 & 74.99 & \textbf{68.30} & 88.35 & 86.10 & \textbf{79.83} & 90.19 & 88.76 & 86.38 \\ \hline 
    \hline
   \textbf{SegVoxelNet} & LiDAR & 25 & 84.19 & \textbf{75.81} & 67.80 & 88.62 & \textbf{86.16} & 78.68 & 90.50 & 88.88 & \textbf{87.34} \\ \hline
   
    \end{tabular}
    \vspace{-3pt}
    \caption{Performance comparison with previous methods on the car class of 3D detection benchmark of KITTI \emph{test} set. Higher scores illustrate more accurate results.}
    \label{test_car}
    \vspace{-5pt}
\end{table*}



\begin{figure*}[t]
  \centering
  \begin{tabular}{cccc}
    \hspace{-0.02\columnwidth}\includegraphics[width=0.505\columnwidth]{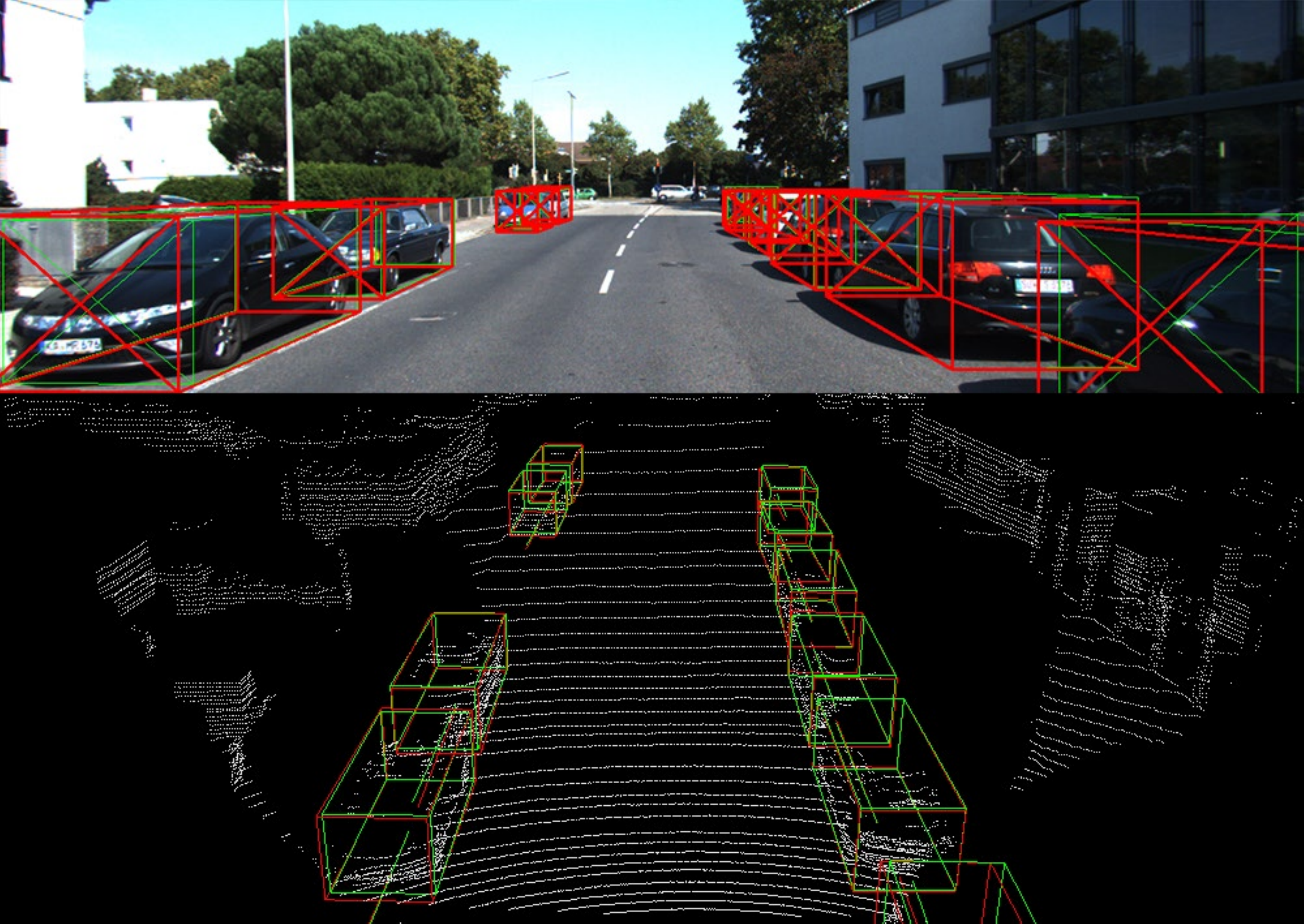} &
    \hspace{-0.04\columnwidth}\includegraphics[width=0.505\columnwidth]{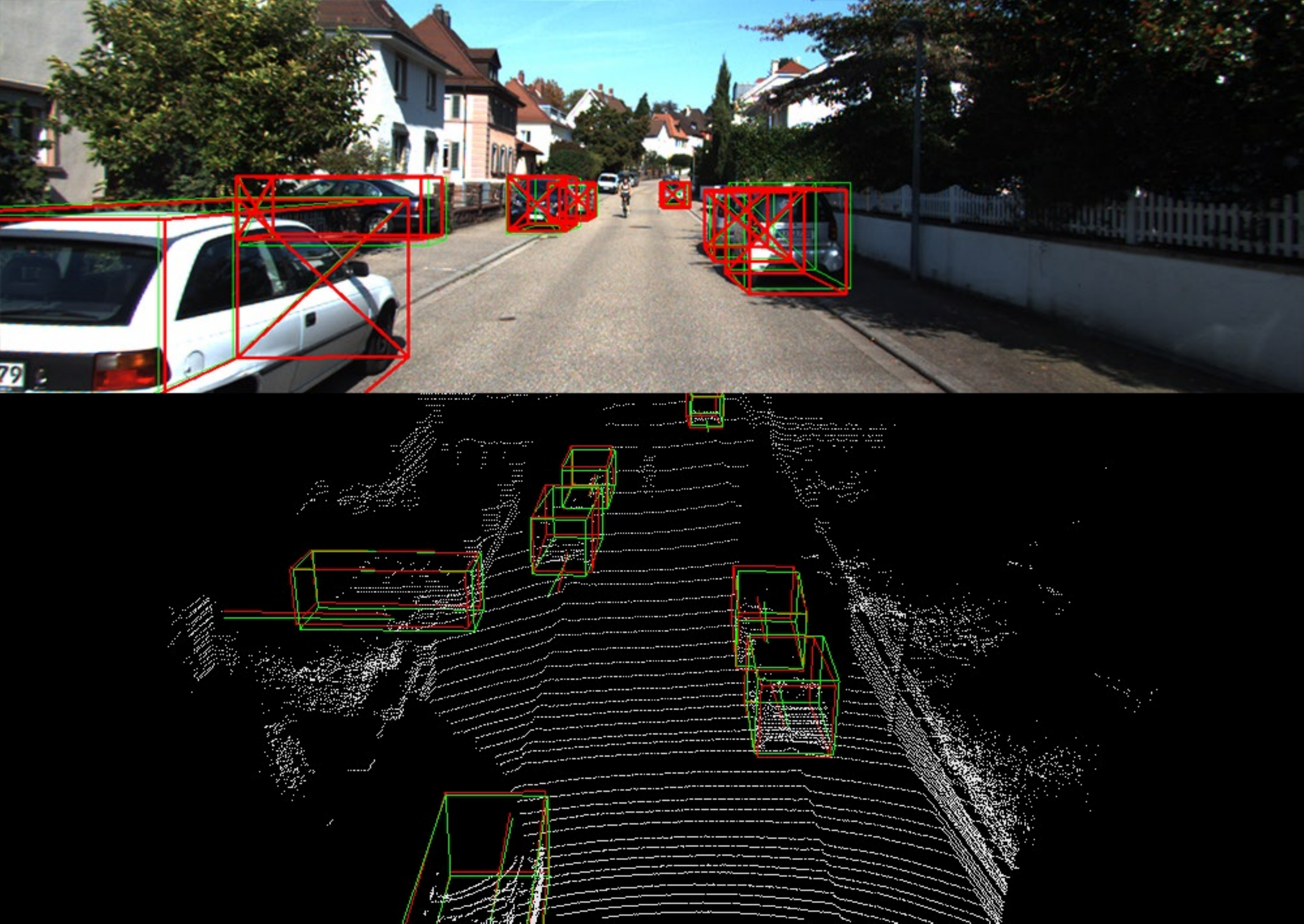}&
    \hspace{-0.04\columnwidth}\includegraphics[width=0.505\columnwidth]{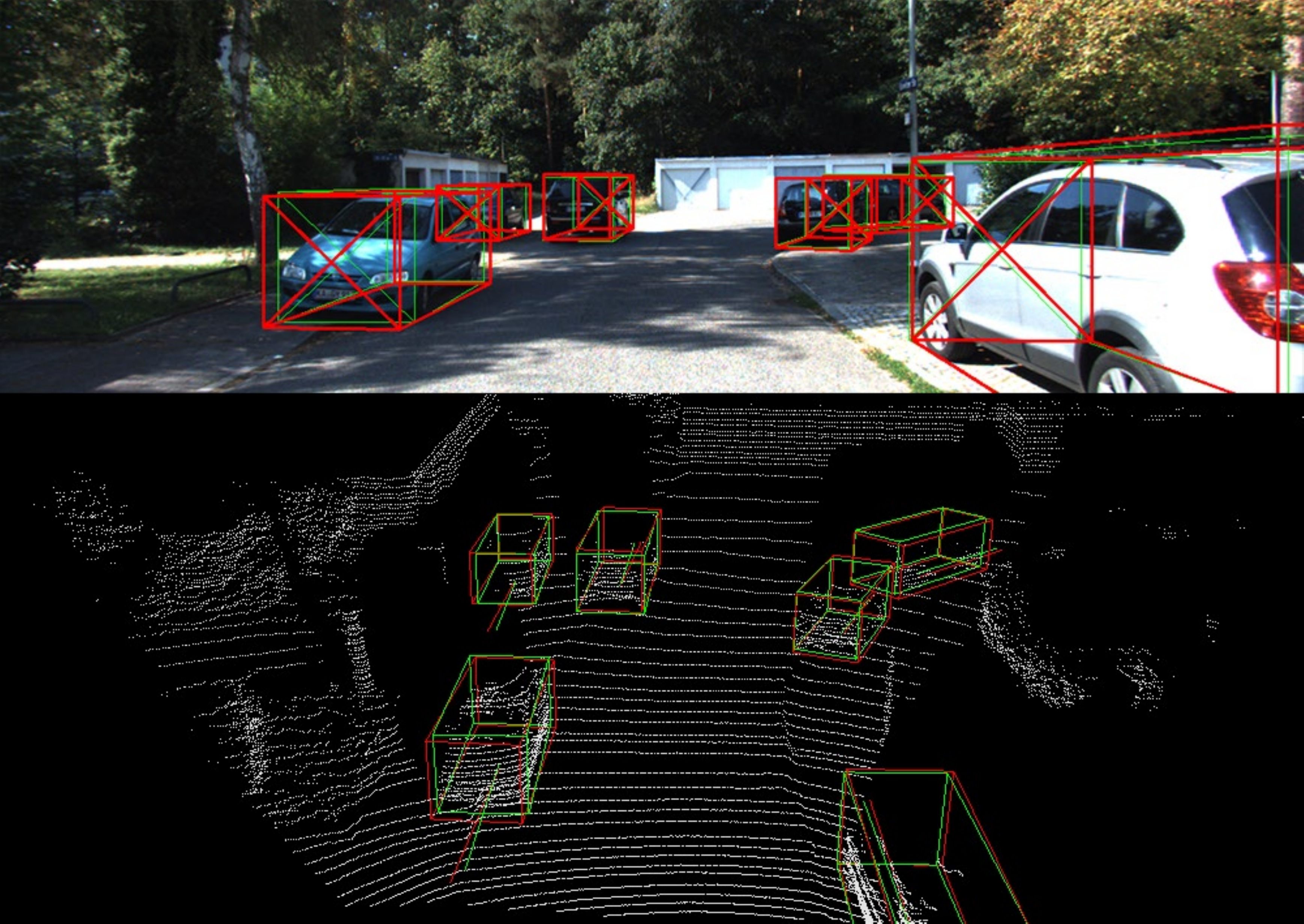} &
    \hspace{-0.04\columnwidth}\includegraphics[width=0.505\columnwidth]{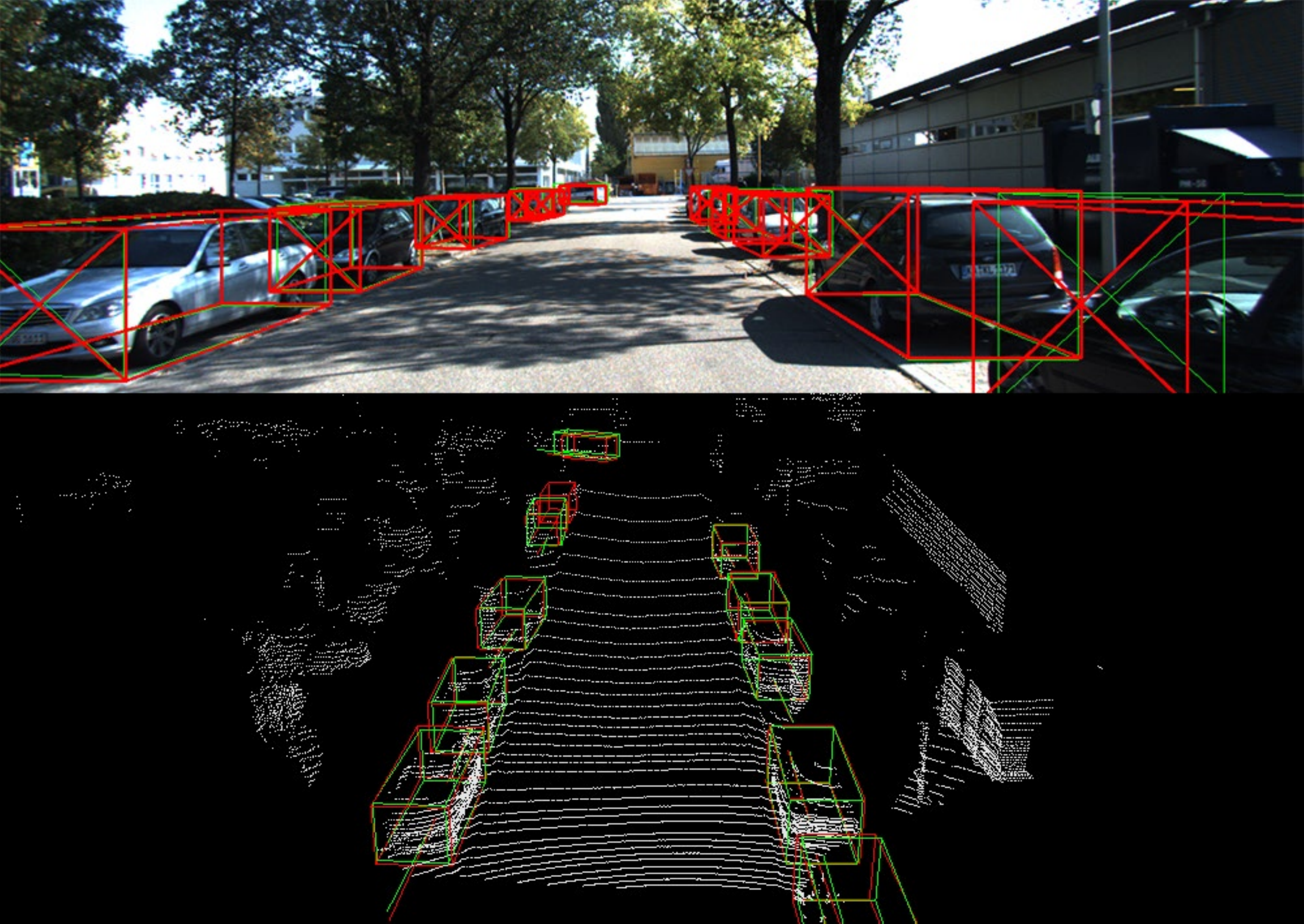}\\
  \end{tabular}
  \vspace{-3pt}
  \caption{Qualitative results of SegVoxel on the KITTI \emph{val} split. 
  The detected vehicles are shown with 3D bounding boxes, and the orientation of each vehicle is illustrated by the outlier middle line on the bottom of the 3D box. Groundtruths are labeled with green while the predictions are labeled with red.}
  \label{qua1}
  \vspace{-5pt}
\end{figure*}

\begin{figure*}
  \centering
  \begin{tabular}{cccc}
    \hspace{-0.02\columnwidth}\includegraphics[width=0.505\columnwidth]{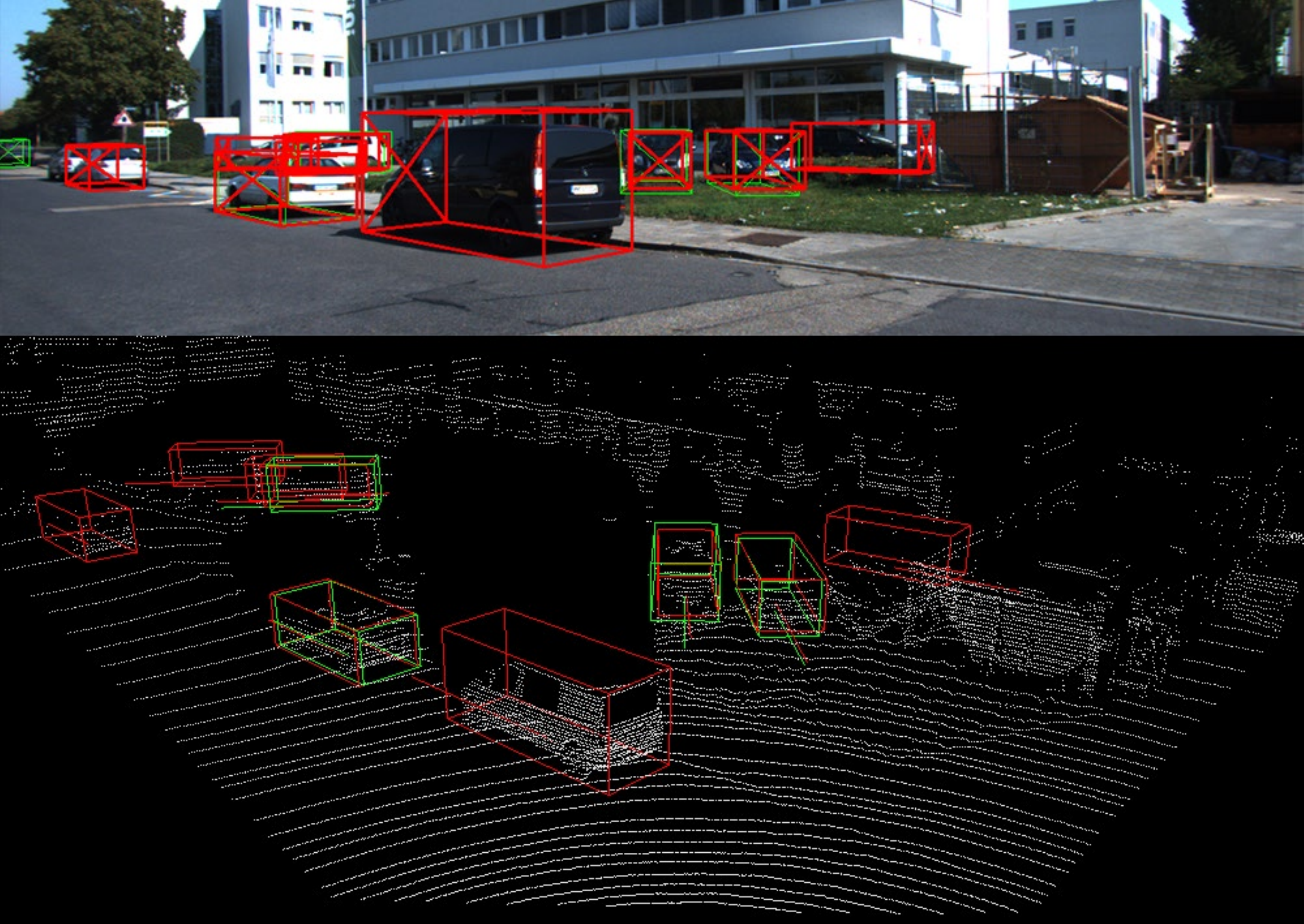} &
    \hspace{-0.04\columnwidth}\includegraphics[width=0.505\columnwidth]{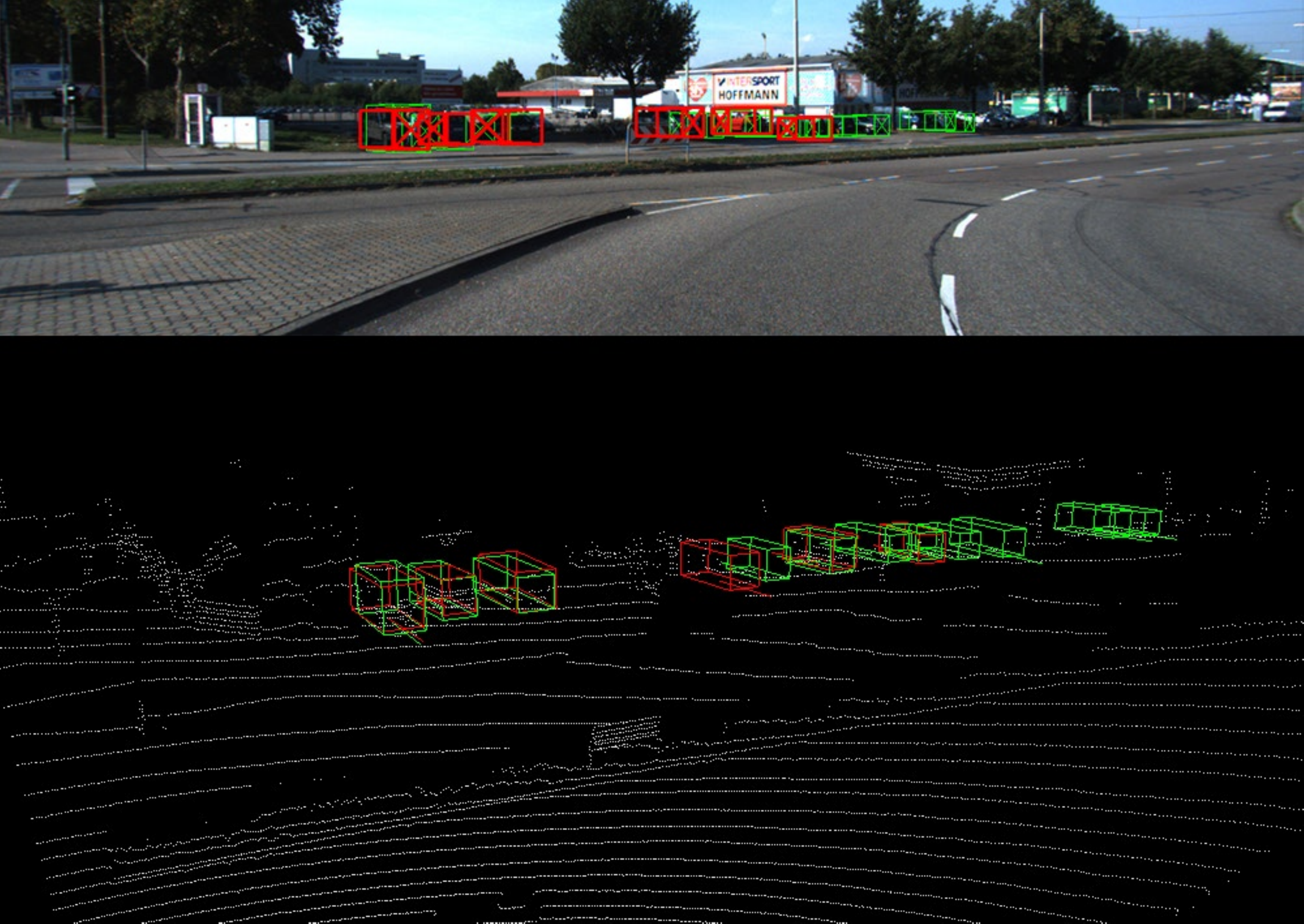} &
    \hspace{-0.04\columnwidth}\includegraphics[width=0.505\columnwidth]{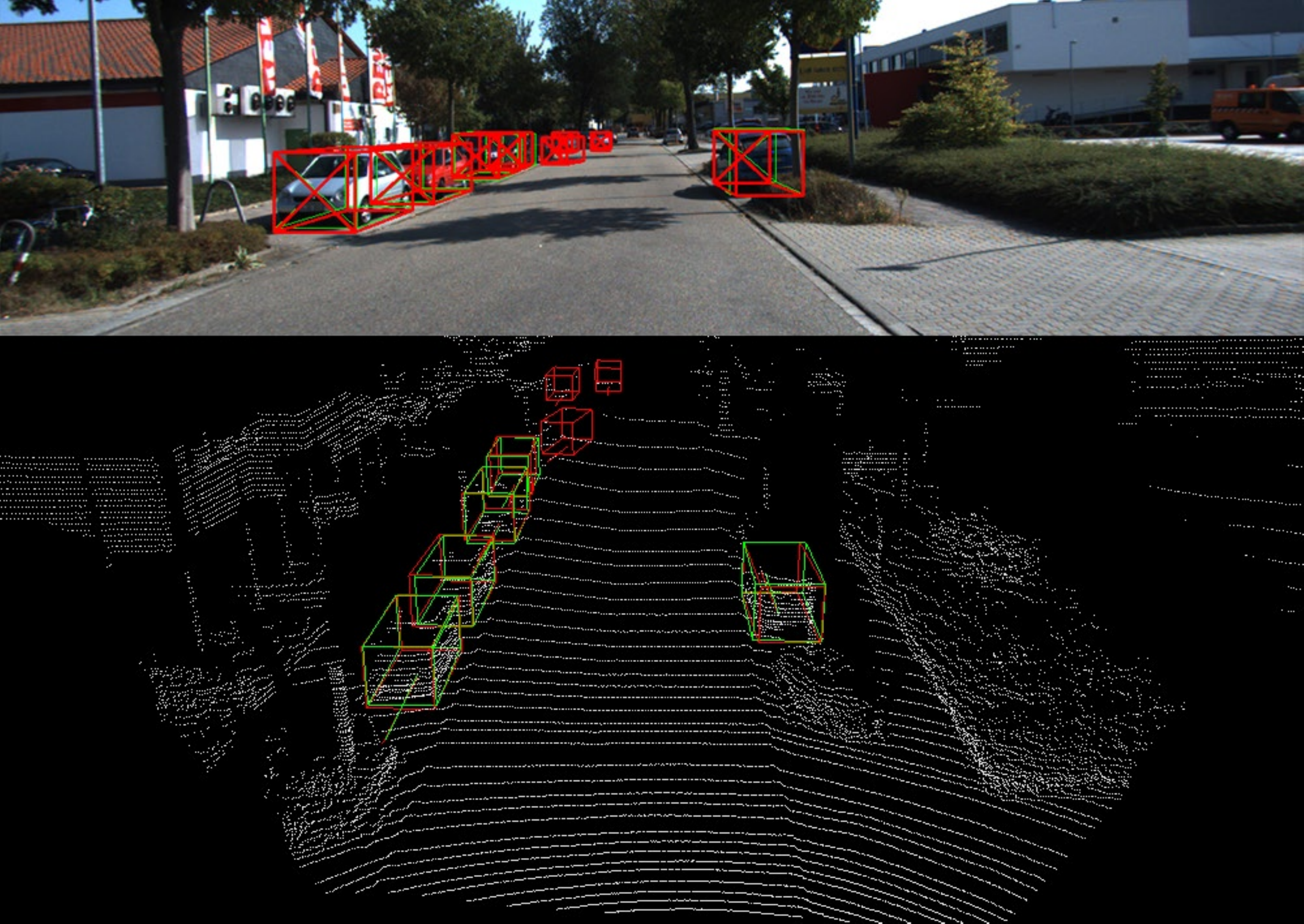} &
    \hspace{-0.04\columnwidth}\includegraphics[width=0.505\columnwidth]{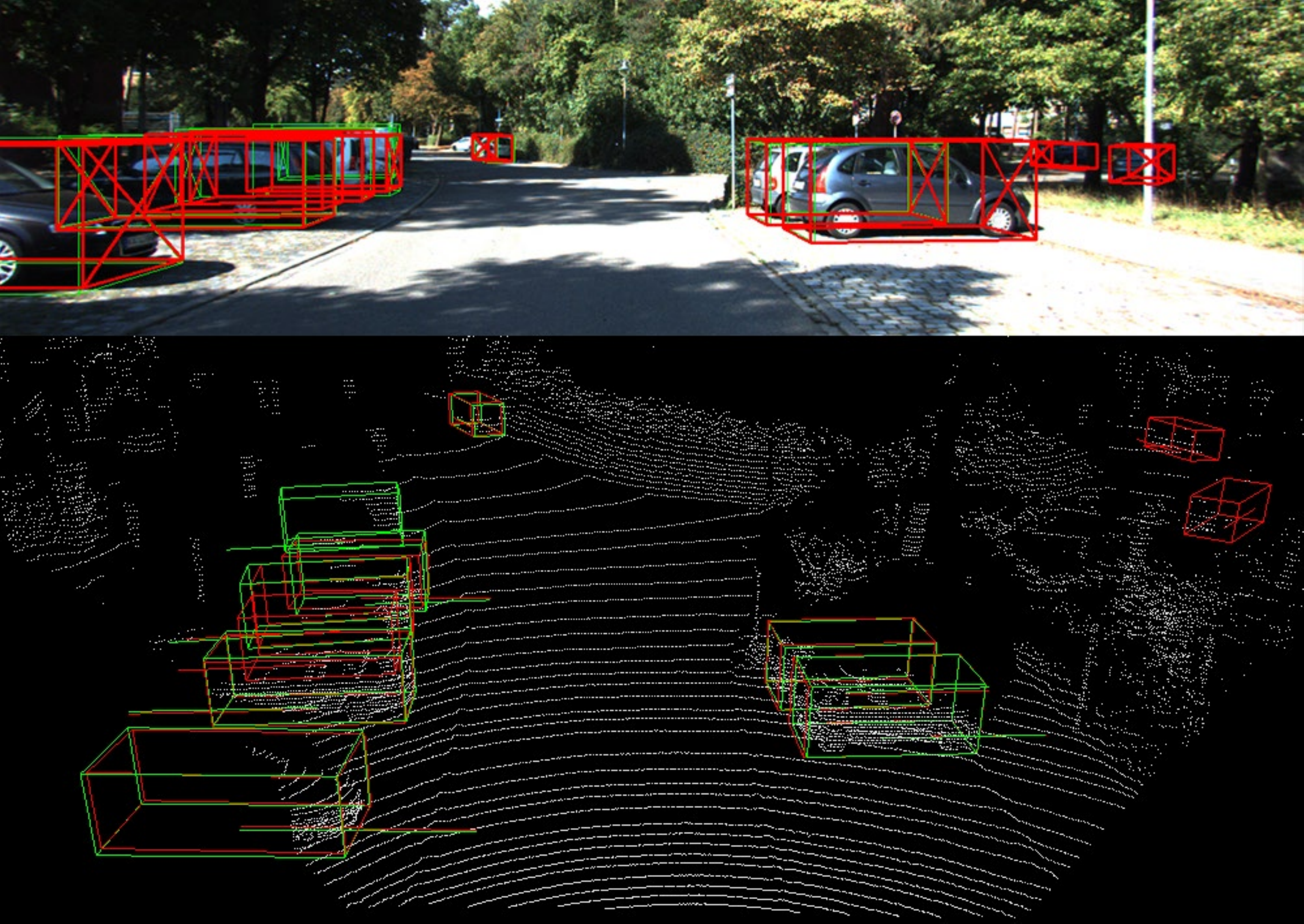} \\
    \vspace{-3pt}
    \small{a)} &
    \small{b)} &
    \small{c)} &
    \small{d)} \\ 
  \end{tabular}
  \vspace{-5pt}
  \caption{Failure cases on KITTI \emph{val} split. In a) Similar class leads to a false positive. In b), c) Far vehicles have only several points, which are quite difficult for 3D vehicle detection. In d) and the others, there all exists miss-labeled ground truths.}
  \label{qua2}
  \vspace{-15pt}
\end{figure*}

\section{Experiments and Results}
In this section, we first introduce our experimental setup including the dataset and implementation details, then we compare with the previous state-of-the-art methods of 3D vehicle detection on both \textsl{val} and \textsl{test} split of KITTI dataset~\cite{Geiger2012CVPR}.

\subsection{Dataset}
The KITTI dataset~\cite{Geiger2012CVPR} is used in all experiments, which contains 7,481 training samples and 7,518 test samples. Each sample consists of a RGB image and a corresponding point cloud. Following common practice, we follow~\cite{chen2017multi} to split the training samples into \textsl{train} split (3,712 samples) and \textsl{val} split (3,769 samples).
We train our model on the \textsl{train} split.
and compare it with other state-of-the-art methods on both \textsl{val} and \textsl{test} split. The KITTI dataset is stratified into easy, moderate and hard difficulties and the leaderboard is ranked by 3D average precision (AP) on moderate difficulty.

\subsection{Implementation Details}
First, we voxelize the point cloud in the range $[0, 70]$, $[-40, 40]$, $[-3, 1]$ with voxel size  $[0.05, 0.05, 0.1]$ along the $x*y*z$ in the LiDAR coordinate system. The maximum number of randomly sampled points in each voxel is set to 5. In our setting, the point cloud from an HDL-64E Velodyne LiDAR has $16k$ non-empty voxels for ~$99.8$\% sparsity in KITTI~\cite{Geiger2012CVPR} dataset. We only store non-empty voxels and their coordinates for reducing memory and further speed-up processing.
The voxel feature encoder (VFE) network consists of four blocks: Block1 (4, 16, 2, 2), Block2 (16, 32, 2, 2), Block3 (32, 64, 3, 2), Block4 (64, 64, 3, 1) with $O=128$ channel output feature maps.
In the semantic context encoder (SCE) network, the multi-scale feature maps are outputted with $2*O$ channels.
In the depth-aware head, we split the feature maps into three parts by $[0, 72]$, $[52, 124]$, $[104, 176]$ in x-axis with two overlapping regions considering the cars in the edge of it.

We train our model initialized as in~\cite{he2015delving}.
For training 3D vehicle detection, a proposal is considered as positive if its 3D IoU with groundtruth boxes is above 0.6 while as negative if its IoU is below 0.45. Each position has two anchors with different orientations. We train SegVoxelNet using the AdamW optimizer~\cite{loshchilov2017fixing} with batch size $6$, weight decay $0.01$ and initial learning rate $2.25*10^{-4}$ for $100$ epochs. 

For data augmentation, we follow SECOND~\cite{yan2018second} to randomly select several ground truths and merge them into the current scene. However, we find out that the ground planes of different scenes have different heights, so we introduce a ground plane equation calculated by RANSAC~\cite{li2017improved} to constrain the augmented samples.
All ground truth boxes and the associated LiDAR points are translated in different axes from $N (0, 0.25)$ and rotated uniformly from $[-\pi/2, \pi/2]$.


\subsection{3D Vehicle Detection Results on KITTI}

\paragraph{Evaluation of 3D vehicle detection} All detection results are measured using the official KITTI evaluation metrics, 
where average precision (AP) is used as an evaluation metric in 3D and BEV by calculating the rotated IoU, and average orientation similarity (AOS) is used as the evaluation metric for orientation estimation, as in~\cite{Geiger2012CVPR}. 

We evaluate the 3D detection results on the KITTI~\cite{Geiger2012CVPR} $test$ dataset in \tref{test_car}. For the most important metric moderate difficulty of 3D mAP, our proposed method SegVoxelNet outperforms all previous methods except~\cite{Liang_2019_CVPR}, which uses both LiDAR and RGB as input and benefits from multi-tasks for 3D vehicle detection. Compared with LiDAR only single stage methods~\cite{voxelnet, lang2018pointpillars, yan2018second}, our method outperforms these with a large margin across all difficulties in 3D vehicle detection. We also achieve comparable results with LiDAR only two-stage method~\cite{shi2019pointrcnn} but have $2.5\times$ speed. For BEV detection, our method achieves better AP on the most important moderate difficulty than all other lidar only methods and obtain comparable AP on easy and hard difficulties. The result of orientation estimation shows that our method predicts much more accurate orientation values in hard difficulty while in easy and moderate difficulties our method achieves better results than other methods except~\cite{shi2019pointrcnn}.

We also report the performance of 3D detection on the $val$ split in \tref{val_car}. Our method outperforms previous state-of-the-art methods in all difficulties, which demonstrated the effectiveness of our method. Besides, our method achieves BEV AP of ($90.37$, $88.13$, $86.94$) and AOS AP of ($90.63$, $89.38$, $88.07$) for the easy, moderate, difficulty respectively.

\paragraph{Qualitative analysis} 
The improvement of integrating semantic segmentation information into detection feature maps can be observed from \fref{segRPN}. We also evaluate different connections between the semantic segmentation feature map and the detection feature map by quantitatively analyzing the different influence in Sec~\ref{ablation_study}. 

We provide qualitative results in \fref{qua1} and \fref{qua2} on the $val$ split of KITTI~\cite{Geiger2012CVPR} dataset.  For ease of interpretation, we visualize the 3D bounding box predictions from both top-view perspective and image front-view perspective. 
\fref{qua1} shows our detection results with tight oriented 3D bounding boxes. 
As shown in \fref{qua2}, although our method predicts some vehicles accurately, there are still some 
difficult examples (several partial occlusion and faraway vehicles or similar classes (vans) which will lead to false positives. Besides, \fref{qua2} shows some false positives which are actual vehicles but not labeled in ground truth annotations.

\subsection{Realtime Inference}
As shown in \tref{test_car}, our method achieves impressive results on 3D vehicle detection with real time efficiency. We divide our SegVoxelNet into different parts and analyze the runtime of each part separately. All runtimes are measured on a desktop with an Intel i7 CPU and 1080ti GPU.
The main inference steps are as follows. First, the point clouds need to be prepossessed as voxels and transferred to GPU ($2.2ms$), then the voxel input tensor is encoded by VFE ($17.8ms$), extracted semantic context encoded feature maps by SCE ($18ms$) and processed by the depth-aware head ($1.8ms$). Finally, NMS is applied on the CPU  ($0.1ms$) for a total runtime of $39.9ms$. Since a LiDAR typically operates at 10HZ and there exists many other speed-up operations, so far, our method can run in real-time efficiency.

\begin{table}[t]
    \centering
    \small
    \begin{tabular}{|c|c|c c c|}
    \hline
    \multirow{2}{*}{Method} & \multirow{2}{*}{Modality} &  \multicolumn{3}{c|}{3D Box} \\ \cline{3-5}
    & & Easy &  \underline{\textbf{Mod.}} & Hard \\ \hline
   \textbf{F-PointNet}~\cite{qi2018frustum} & LiDAR \& RGB & 83.76 & 70.92 & 63.65 \\ \hline
   \textbf{AVOD-FPN}~\cite{ku2018joint} & LiDAR \& RGB  & 84.41 & 74.44 & 68.65 \\ \hline\hline 
   \textbf{PointRCNN}~\cite{shi2019pointrcnn} & LiDAR \tiny{(Two Stage)} & 88.88 & 78.63 & 77.38 \\ \hline\hline 
   \textbf{VoxelNet}~\cite{voxelnet} & LiDAR & 81.98 & 65.46 & 62.85 \\ \hline
   \textbf{SECOND}~\cite{yan2018second} & LiDAR & 87.43 & 76.48 & 69.10 \\ \hline
   \textbf{PointPillars}~\cite{lang2018pointpillars} & LiDAR & - & 77.98 & -\\ \hline\hline 
   \textbf{SegVoxelNet} & LiDAR & \textbf{89.35} & \textbf{79.05} & \textbf{77.41} \\ \hline
    \end{tabular}
    \caption{Performance comparison with previous methods on the car class of KITTI \textsl{val} 3D detection benchmark.}
    \label{val_car}
    \vspace{-5pt}
\end{table}

\begin{table}[t]
    \centering
    \small
    \begin{tabular}{c c | c |ccc}
    \hline
      \multicolumn{2}{c|}{Seg} & \multirow{2}{*}{Depth-aware} &  \multirow{2}{*}{$AP_{E}$} & \multirow{2}{*}{$AP_{M}$} & \multirow{2}{*}{$AP_{H}$}  \\ \cline{1-2} 
      \scriptsize{re-weight} & \scriptsize{concat} & & & & \\ 
      \hline
       &  & & 88.55& 78.32& 76.71 \\ 
      \hline
      \checkmark&  & & $+0.60$ & $+0.47$ &$+0.48$ \\ 
      &\checkmark  & & $+0.62$ &$+0.43$ &$+0.52$\\ 
      & & \checkmark & $+0.28$ &$+0.37$ &$+0.40$\\
      \checkmark& & \checkmark & $\textbf{+0.80}$ & $\textbf{+0.73}$ & $\textbf{+0.70}$\\ 
    \hline
    \end{tabular}
    \caption{Performance for different fusion between semantic segmentation feature map and detection feature map and our designed depth-aware head. $AP_{E}$, $AP_{M}$, $AP_{H}$ denote the average precision of 3D box on KITTI $val$ split for easy, moderate, hard difficulty separately.}
    \label{ablation}
    \vspace{-6pt}
\end{table}

\begin{table}
    \centering
    \small
    \begin{tabular}{c|c c c}
    \hline
      Depth-aware &  $mAP_N$&  $mAP_M$& $mAP_F$ \\
      \hline
       & 89.01 & 51.92 & \textbf{10.72} \\ 
      \checkmark & \textbf{89.33} & \textbf{53.24} & 10.06 \\ \hline
      \checkmark & $+0.32$ & $+1.32$ & $-0.66$ \\ \hline
    \end{tabular}
    \caption{Performance for different branch on KITTI $val$ dataset for the whole difficulty strata. $mAP_N$, $mAP_M$, $mAP_F$ denote the mean average precision (mAP) of 3D box for the whole difficulty strata in the Near (N), Middle (M) and Far (F) branch respectively.}
    \label{branch}
    \vspace{-5pt}
\end{table}

\subsection{Ablation Study} \label{ablation_study}
In this section, we conduct extensive ablation experiments to analyze the effectiveness of different components of SegVoxelNet. All experiments are trained on $train$ split and evaluated on the $val$ split with the car class.

\textbf{Effect of different fusion between semantic segmentation feature maps and detection feature maps:}
In Semantic Context Encoder(SCE), Semantic context information is introduced to highlight the existing vehicle region and suppress the background. We further evaluate the quantitative effects of the proposed fusion methods. We compare the proposed  attention residual learning  with simple concatenation along the channel axis. \tref{ablation} shows the 3D vehicle detection performance with re-weight and concatenation. It can be seen that these both methods are better than the baseline method without the semantic branch, which proves the effectiveness of the semantic context information for detection. While the performance gap between these two variants is very small, the proposed re-weight method results in fewer parameters and is more efficient than the concatenation alternative. 

\textbf{Effect of depth-aware head:}
In Sec. \ref{detectionHead}, for high efficiency, we design a depth-aware head with small overlapping regions between different parts on the SEC output feature map, which leads to neglected computation. To show the importance of the depth-aware head, we compare it with our baseline method on 3D vehicle detection performance as shown in \tref{ablation}. The depth-aware head improves the 3D AP with about 0.3 $\sim$ 0.4 gains through three difficulties $AP_{E}$, $AP_{M}$, $AP_{H}$.
The gain of each branch is shown in \tref{branch}. We achieve a large margin improvement of 3D vehicle detection in the Middle (M) branch and a small improvement in the Near (N) branch but a small decrease in the Far (F) branch. We think it may be caused by the unlabeled vehicles in the far distance, as shown in \fref{qua2}.

\section{Conclusion}
In this paper, we propose a unified framework called SegVoxelNet that incorporates semantic information into 3D vehicle detection, where semantic context becomes active guidance for 3D vehicle detection. Moreover, an efficient depth-aware head is designed for vehicles with different depths in autonomous driving scenarios. Compared with other LiDAR-only methods, the experiments show that our SegVoxelNet achieves state-of-the-art results on the challenging 3D detection benchmark of KITTI dataset~\cite{Geiger2012CVPR} with real-time efficiency.

\section*{ACKNOWLEDGMENT}
This project was supported by the National Key R\&D Program of China (No.2017YFB1002700, No.2017YFB0203000) and NSFC of China (No.61632003, No.61661146002, No.61631001).

\bibliographystyle{IEEEtran}
\bibliography{}







\end{document}